\documentclass[runningheads,a4paper, UTF8]{llncs}
\usepackage{CJKutf8}
\usepackage[utf8]{inputenc}
\usepackage{multirow}
\usepackage{amssymb}
\usepackage{amsmath}
\usepackage{xfrac}
\usepackage{graphicx}
\usepackage[table,x11names]{xcolor}
\usepackage{url}
\makeatletter
\newcommand*\bigcdot{\mathpalette\bigcdot@{.5}}
\newcommand*\bigcdot@[2]{\mathbin{\vcenter{\hbox{\scalebox{#2}{$\m@th#1\bullet$}}}}}
\makeatother

\usepackage[misc]{ifsym}

\usepackage{hyperref}
\hypersetup{
  pdfinfo={
  pdfproducer={},
  Title={},
  Subject={},
  Author={},
  }
}

\begin{document}

\mainmatter  % start of an individual contribution

\title{FIRE: Unsupervised bi-directional inter-modality registration using deep networks}
%\titlerunning{Lecture Notes in Computer Science: Authors' Instructions}


\author{Chengjia Wang $ ^{1(\textrm{\Letter})} $ \and Giorgos Papanastasiou $^{1}$ \and Agisilaos Chartsias $^{2}$ \and Grzegorz Jacenkow $^{2}$ \and Sotirios A. Tsaftaris $^{2}$ \and Heye Zhang $ ^{3} $}
%
%\author{************* $ ^{1(\textrm{\Letter})} $ \and ********************* $^{1}$ \and ******************* $^{2}$ \and ***************** $^{2}$ \and ************ ********* $^{2}$ \and ********** $ ^{3} $}

% authors: Chengjia, Giorgos, Agis, Greg, Sotos


\institute{$^{1}$Edinburgh Imaging Facility QMRI, Centre for Cardiovascular Science, University of Edinburgh, Edinburgh EH16 4TJ, UK \\ \url{chengjia.wang@ed.ac.uk} \\
$^{2}$Institute for Digital Communications, School of Engineering, University of Edinburgh, West Mains Rd, Edinburgh EH9 3FB, UK\\
$^{3}$School of Biomedical Engineering, Sun Yat-Sen University, Guangming, Shenzhen 518055, China\\
}

%\institute{$^{1}$******************************* ************************************************ \\ \url{**********************} \\
%$^{2}$********************************************** **************************************************** *********************\\
%$^{3}$********************************************** **********************************************\\
%}

%\institute{$^{1}$**************************************************** ************************ \\
%\url{**********************} \\
%$^{2}$*********************************************************
%**************\\
%$^{3}$********************************************************** ***********\\
%}

% (search for the string "" where a contribution starts).

\toctitle{Lecture Notes in Computer Science}
\tocauthor{Authors' Instructions}
\maketitle

\begin{abstract}
Inter-modality image registration is an critical preprocessing step for many applications within the routine clinical pathway. This paper presents an unsupervised deep inter-modality registration network that can learn the optimal affine and non-rigid transformations simultaneously. Inverse-consistency is an important property commonly ignored in recent deep learning based inter-modality registration algorithms. We address this issue through the proposed multi-task architecture and the new comprehensive transformation network. Specifically, the proposed model learns a modality-independent latent representation to perform cycle-consistent cross-modality synthesis, and use an inverse-consistent loss to learn a pair of transformations to align the synthesized image with the target. We name this proposed framework as FIRE due to the shape of its structure. Our method shows comparable and better performances with the popular baseline method in experiments on multi-sequence brain MR data and intra-modality 4D cardiac Cine-MR data. 
\end{abstract}

\section{Introduction}
Modern medical diagnosis benefits from fusion complementary information obtained by different modalities. This makes inter-modality image registration an critical pre-processing task for many applications within the routine clinical pathway \cite{Rueckert2010}. (In this paper use ``modality'' to uniformly address data acquired with different imaging techniques and with different parametric setups.) Traditional and early learning-based methods typically model the registration problem as an iterative optimization process to find an optimal value of manually designed similarity metrics, thus they are often computationally expensive  \cite{Haskins2019}. Furthermore, manually designed metrics have limited robustness and performances upon registering inter-modality data. 

As discussed in \cite{Haskins2019}, in the passed decade, a variety of deep learning based methods have been proposed that can predict the geometric correspondences between a pair of images in one pass. But most existing methods are based on supervised learning and requires manually generated ground truths, such as, pre-aligned image pairs, simulated transformation fields and segmentation labels \cite{Cao2017}\cite{Krebs2017}\cite{Rohe2017}. At the same time, present unsupervised methods \cite{Jaderberg2015} have been mostly tested only on limited subsets of 3D volumes or 2D slices with small misalignments and require reliable affine registration as a preprocessing. Deep learning models that can perform both affine and non-rigid registrations \cite{Vos2019} often requires to apply two independent models for both types of transformation. In this paper, we presents an unsupervised deep inter-modality registration network that can learn the optimal affine and non-rigid transformations simultaneously.

\begin{figure}[!t]
\centering
\includegraphics[width=0.8\linewidth]{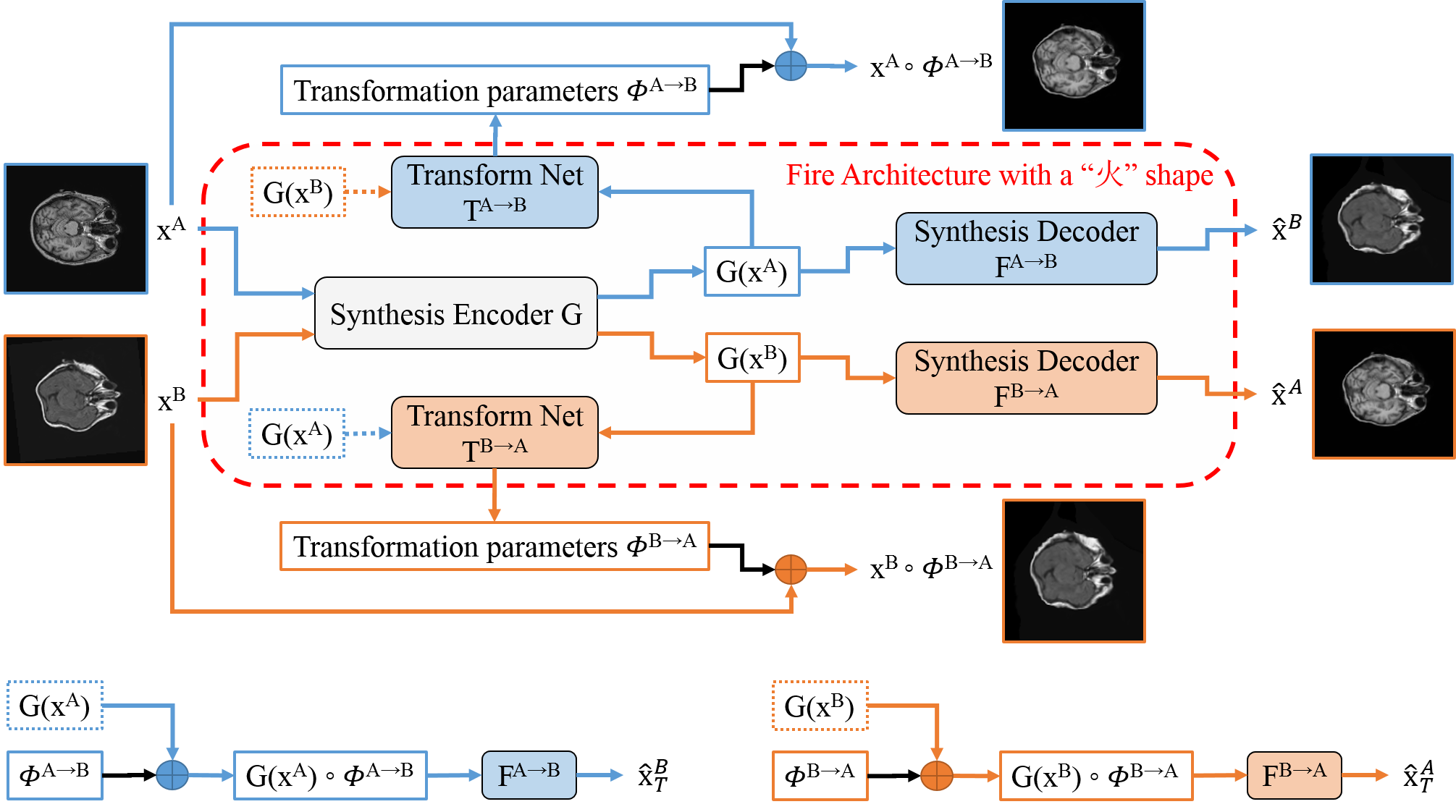}
\caption{Architecture of the FIRE model: a synthesis encoder, $G$, that extracts modality-independent features; two synthesis decoders, $ F^{A\rightarrow B} $ and $ F^{B\rightarrow A} $, that map the features extracted by $G$ to synthesized images; and two transformation networks, $ T^{A\rightarrow B} $ and $ T^{B\rightarrow A} $, that predict the transformation fields.}
\label{fig:FIREArc}
\end{figure}

Our method solves $n$-D image registration problems through cross-modality image synthesis and inverse-consistent transformations \cite{Christensen2001}. The cycle consistency adversarial loss has been widely used within this type of methods. Similarly, inverse-consistency (or bi-directional) transformation has been a favorable property for better maintenance of the neighbourhood topology and anatomy of organs. However, most previous works failed to address this issue and solely estimate asymmetric transformations. Two inverse-consistent models presented in concurrent preliminary works \cite{Qin2019}\cite{Zhang2018} are closer to the proposed method. However, \cite{Zhang2018} is for intra-modality registration, and \cite{Qin2019} has only been tested for 2D non-rigid registration.

We named the proposed model as \textit{FIRE} because its architecture, as shown in Fig. \ref{fig:FIREArc}, display a shape of the character ``\begin{CJK*}{UTF8}{gbsn}火\end{CJK*}'' (a component representing fire in Chinese language). We present experiments demonstrating that our method achieves state-of-the-art performances registering multi-sequence brain MR data with aggressive simulated deformations and intra-modality 4D cardiac MR data. To sum up, contributions of this paper include: (1) the ``\begin{CJK*}{UTF8}{gbsn}火\end{CJK*}''-shape FIRE architecture for inverse-consistent inter-modality registration; (2)simultaneous learning for affine and non-rigid transformation; and (3) new regularization for non-rigid registration using the predicted affine transformation. 

\section{Method} 
With two images $x^A$ and $x^B$, the proposed FIRE model predicts two transformations $ \phi ^{A\rightarrow B} $ and $\phi ^{B\rightarrow A} $ to warp the images into $ x^A \circ \phi ^{A\rightarrow B} $ and $ x^B \circ \phi ^{B\rightarrow A} $. Transformation fields are obtained by minimizing a loss $ \mathcal{L}\left( x^A, x^B, \phi ^{A\rightarrow B}, \phi ^{B\rightarrow A} \right) $ (or $\mathcal{L}$ for clear and effective explanation in this paper). Computations described in this section are based on input data normalized to the range $[-1, 1]$.

\subsection{Architecture}
The FIRE model consists of five sub-networks (Fig. \ref{fig:FIREArc}): a synthesis encoder, $G$, that extracts modality-independent features $G(x^A)$ and $G(x^B)$; two synthesis decoders, $ F^{A\rightarrow B} $ and $ F^{B\rightarrow A} $, that map the features extracted by $G$ to synthesized images $ \hat{x}^B = F^{A\rightarrow B}(G(x^A))$ and $ \hat{x}^A = F^{B\rightarrow A}(G(x^B)) $; and two transformation networks, $ T^{A\rightarrow B} $ and $ T^{B\rightarrow A} $, that predict the transformation fields $ \phi^{A\rightarrow B} = T^{A\rightarrow B}(G(x^A), G(x^B)) $ and $ \phi^{B\rightarrow A} = T^{B\rightarrow A}(G(x^B), G(x^A)) $. In the training stage, $ G(x^A) $ and $ G(x^B) $ are also warped into $ G(x^A) \circ \phi^{A\rightarrow B} $ and $ G(x^B) \circ \phi^{B\rightarrow A} $, then used to generate synthesized images,  $ \hat{x}^{B}_{T} = F^{A\rightarrow B}(G(x^A) \circ \phi^{A\rightarrow B}) $ and $ \hat{x}^A_T = F^{B\rightarrow A}(G(x^B) \circ \phi^{B\rightarrow A})$. 

\subsubsection{Synthesis Encoder and Decoder}
\begin{figure}[!t]
\centering
\includegraphics[width=0.8\linewidth]{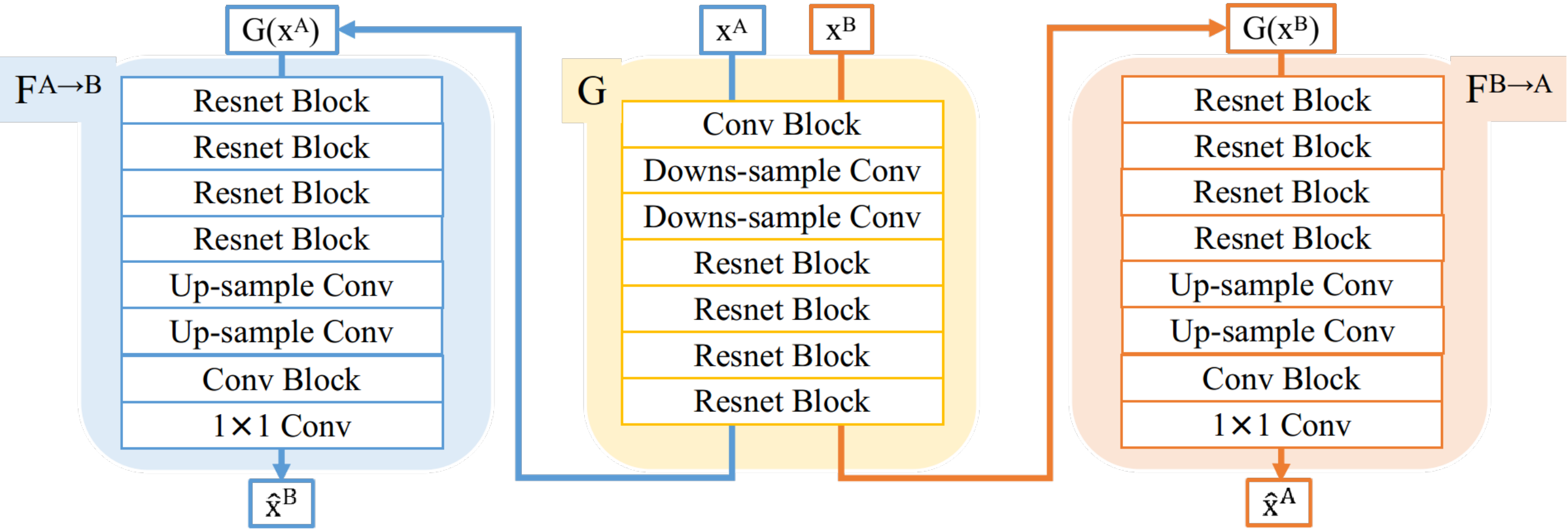}
\caption{Architecture of the synthesis encoder and decoders.}
\label{fig:SynNets}
\end{figure}

Fig. \ref{fig:SynNets} shows the details about architecture of the synthesis networks. The encoder $ G $ contains a input convolutional layer, two downsample convolutional layers and four Resnet blocks. A decoder network starts with four Resnet blocks, followed by two upsample convolutional layers, and output convolutional layers. All convolutional layers use a kernel size of 3, followed by an instance normalization layer. 

\subsubsection{Transformation Network}
\begin{figure}[!t]
\centering
\includegraphics[width=0.9\linewidth, height=0.45\linewidth]{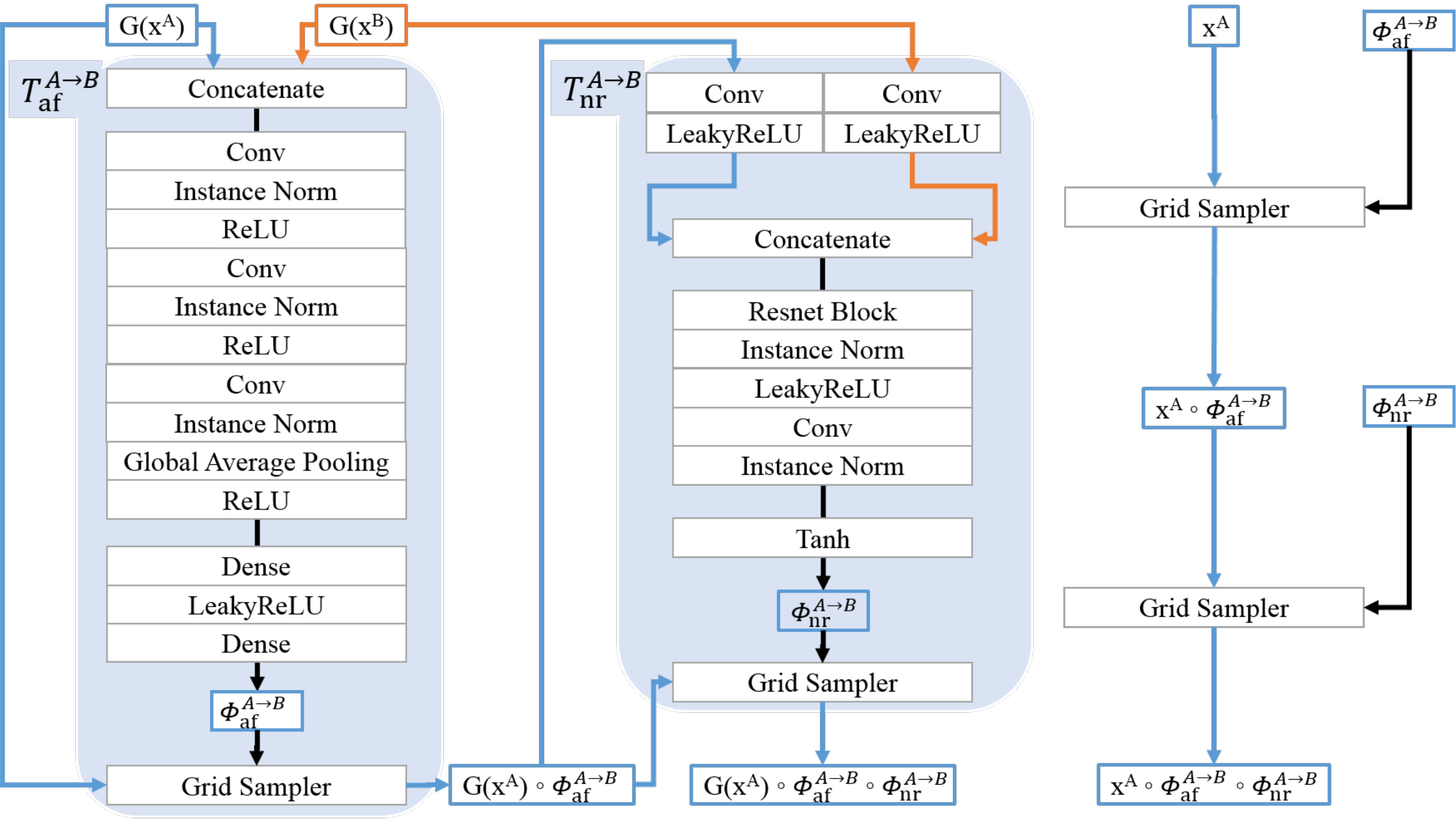}
\caption{Architecture of the transformation networks.}
\label{fig:STNNets}
\end{figure}

A transformation network $T^{\bigcdot \rightarrow \bigcdot}$ learns both an affine transformation $\phi_{af}$ and a non-rigid transformation $ \phi_{nr} $ given $G(x^A)$ and $ G(x^B) $. Fig. \ref{fig:STNNets} presents the architecture of $ T^{A\rightarrow B} $ and $ T^{B\rightarrow A} $ has the same architecture. The affine transformation sub-network $ T_{af} $ has a similar structure of the original spatial transformation networks (STN). A global average pooling layer is used to resample conv features into a fixed size feature vector. Affine transformation is calculated using two fully connected layers. The non-rigid transformation sub-net $ T_{nr}^{A\rightarrow B} $ takes $ G(x^A) \circ T^{A\rightarrow B}_{af} $ and $ G(x^B) $ as input, and process them parallel layers first. Extracted features are then concatenated to produce the non-rigid deformation $\phi_{nr}^{A\rightarrow B}$. The last $Tanh$ layer is for a normalized coordinate system where a coordinate $\mathbf{p} \in [-1, 1]^n$ on a $n$-D image.

\subsection{Loss Functions and Training Procedure}
The $A\rightarrow B$ synthesis generate two synthesized images $ \hat{x}^B $ and $ \hat{x}^{B}_T $, where $ \hat{x}^{B} $ is aligned with $ x^A $ and $ \hat{x}^{B}_T $ is identical to the target image $ x^B $. The backward $B\rightarrow A$ synthesis and registration are performed through the same pipeline using the ``$B\rightarrow A$'' networks. Losses used for training FIRE model includes a synthesis loss, $ \mathcal{L}_{syn} $, and a registration loss, $ \mathcal{L}_{reg} $. A new regularization $ \mathcal{R} $ is used for spatially smooth and topology-preserving deformation. The loss function of the proposed FIRE model is defined as:
\begin{equation}
\mathcal{L} = \mathcal{L}_{syn} + \mathcal{L}_{reg} + \mathcal{R}.
\label{equ:FIRELoss}
\end{equation}

\subsubsection{Synthesis Loss}
The synthesis loss includes four terms for different purposes. First, for accurate cross-domain synthesis, we define a synthesis accuracy loss $ \mathcal{L}_{syn, acc} = RMS(\hat{x}^{B}_T,  x^{B}) + RMS(\hat{x}^{A}_T, x^{A}) $ using the root-mean-square (RMS) error. Second, $ G $ is expected to extract modality-independent features, thus features extracted from aligned image pairs should be identical regardless their modalities. So we define a feature loss $ \mathcal{L}_{syn, fea} = RMS(G(x^A), G(x^B) \circ \phi^{B\rightarrow A}) + RMS(G(x^B), G(x^A)\circ \phi^{A\rightarrow B}) $. The third cycle-consistency loss is defined as $ \mathcal{L}_{syn,cyc} = RMS(F^{B\rightarrow A}(G(\hat{x}^B)), x^A) + RMS(F^{A\rightarrow B}(G(\hat{x}^A)), x^B) $ for robust cross-modality synthesis. Finally, for alignment between $ x^{\bigcdot} $ and $ \hat{x}^{\bigcdot} $ we define a alignment loss $ \mathcal{L}_{syn, align} = RMS(G(x^A), G(\hat{x}^B)) + RMS(G(x^B), G(\hat{x}^A)) $. To sum up, the FIRE synthesis loss is:
\begin{equation}
\mathcal{L}_{syn} = \mathcal{L}_{syn, acc} + \mathcal{L}_{syn, fea} + \mathcal{L}_{syn, cyc} + \mathcal{L}_{syn, align}.
\label{equ:FIRESynLoss}
\end{equation}

\subsubsection{Registration Loss}
Transforming the features extracted by $ G $ is for synthesis purpose and registration is achieved by applying the transformations $ \phi^{\bigcdot \rightarrow \bigcdot} = \phi^{\bigcdot \rightarrow bigcdot}_{af} \circ \phi^{ \bigcdot \rightarrow \bigcdot}_{nr}$ to input images. Here we define a registration accuracy loss $ \mathcal{L}_{reg, acc} = RMS(F^{A\rightarrow B}(G(x^A \circ \phi^{A\rightarrow B})), x^B) + RMS( F^{B\rightarrow A}(G(x^B \circ \phi^{B\rightarrow A})) ) $. For mutually inversed transformations $ \phi^{A\rightarrow B} $ and $ \phi^{B\rightarrow A} $, we define a inverse-consistency loss $ \mathcal{L}_{reg, ic} = RMS( x^A, x^A \circ \phi^{A\rightarrow B} \circ \phi^{B\rightarrow A} ) + RMS( x^B, x^B \circ \phi^{B\rightarrow A} \circ \phi^{A\rightarrow B} ) $. The registration loss is computed as:
\begin{equation}
\mathcal{L}_{reg} = \mathcal{L}_{reg, acc} + \mathcal{L}_{reg, ic}.
\label{equ:RegLoss}
\end{equation}
%In practice, the non-rigid transformation field $ \phi^{\bigcdot}_{nr} $ is calculated on $ G(x^{\bigcdot}) $, and resampled linearly before applied to the image $ x^{\bigcdot} $. 

\subsubsection{Regularization}
Previous works regularize the non-rigid transformation fields by a smoothness regularization $ \mathcal{R}_{smooth} = $ $ \parallel \nabla ^{2} \phi^{A\rightarrow B}_{nr} \parallel ^2 + \parallel \nabla ^{2} \phi^{B\rightarrow A}_{nr} \parallel ^2$ where $ \nabla $ is the Laplacian operator. In this work, the estimated affine transformations should keep the non-rigid transformations in the minimal level. In the synthesis process, the affinely transformed features, $ G(x^A) \circ \phi^{A\rightarrow B}_{af} $ and $ G(x^B) \circ \phi^{B\rightarrow A}_{af} $, can be input into the synthesis decoders to obtain $ F^{A\rightarrow B}(G(x^A) \circ \phi^{A\rightarrow B}_{af}) $ and $ F^{B\rightarrow A}(G(x^B) \circ \phi^{B\rightarrow A}_{af}) $. The regularization of synthesis is then computed as $ \mathcal{R}_{syn} = RMS(x^B,F^{A\rightarrow B}(G(x^A) \circ \phi^{A\rightarrow B}_{af})) + RMS( x^A, F^{B\rightarrow A}(G(x^B) \circ \phi^{B\rightarrow A}_{af}) )$. Similarly, a regularization of registration, $ \mathcal{R}_{reg}$, is computed as:
\begin{center}
$RMS(x^B, F^{A\rightarrow B}(G(x^A \circ \phi^{A\rightarrow B}_{af}))) + RMS(x^A, F^{B\rightarrow A}(G(x^B \circ \phi^{B\rightarrow A}_{af})))$. 
\end{center}
To sum up, the regularization of the proposed FIRE model is:
\begin{equation}
\mathcal{R} = \mathcal{R}_{syn} + \mathcal{R}_{reg} + \lambda \mathcal{R}_{smooth},
\label{equ:FIRERegularization}
\end{equation}
where $ \lambda $ is a scaling parameter of $ \mathcal{R}_{smooth} $. Empirically, when registering $n$-D images, and $G$ has $C_{G}$ output channels, $\lambda = \sfrac{2^{2n}}{10N}$, where $N$ represents number of points in an input image.

\subsubsection{Optimization}
We use three Adam optimizers to update parameters of $ T^{\bigcdot \rightarrow \bigcdot}_{af} $, $ T^{\bigcdot \rightarrow \bigcdot}_{nr} $ and the rest networks separately in each three consecutive iterations for a stable convergence. Learning rates for training $ T^{\bigcdot \rightarrow \bigcdot}_{af} $ and $ T^{\bigcdot \rightarrow \bigcdot}_{nr} $ are set to $ 5 \times 10^{-5} $, and to $ 10^{-4} $ for training $ G $ and $F^{\bigcdot \rightarrow \bigcdot}$. 

\section{Experiments}
\subsubsection{MRBrainS}
We use a dataset of 3T multi-sequence brain MR data by mixing the training data set from the MRBrains18 \footnote{https://mrbrains18.isi.uu.nl/} and MRBrains13 \footnote{http://mrbrains13.isi.uu.nl/} Challenges. The dataset contains co-registered 3D T1-weighted, inversion recovery (IR) and T2-FLAIR data acquired from 12 subjects. All scans have a voxel size of $0.958 \times 0.958 \times 3.0 mm^{3}$. We use manual segmentations of 3 anatomical structures to evaluate performances of registration algorithms. Data from 8 patients are for training, 1 for validation and 3 for testing. For both 3D and 2D registration, we resampled all data to $ 1.28mm^3 $ per voxel. We perform 2D and 3D registration between T1 and FLAIR data, and 2D registration between IR and FLAIR data. In the training stage, randomly generated affine and non-rigid transformation are applied to the moving image.

\subsubsection{ACDC}
For intra-modality registration, we use 4D cine-MR data from the 2017 ACDC Challenge \footnote{https://www.creatis.insa-lyon.fr/Challenge/acdc}. The training dataset includes data from 100 patients with a variety of pathology. The in-plane resolution is between $1.37$ and $1.68 mm^2/pixel$, and each 4D image has 28 to 40 phases that cover completely or partially the cardiac cycle. Manual segmentation of 2 phases are provided for each 4D data. We use all phases for training and the two segmented phases for testing. We use 40 patients for training, 10 for validation, and 50 for testing.

\subsubsection{Evaluation Metrics and Baselines}
We evaluate our method using the overlap of the segmented objects measured by Dice metric. Higher Dice scores indicate better registration performances. Previous comparison studies show that Symmetric Normalization (SyN) \cite{Avants2008} implemented in the ANTs toolbox \footnote{http://stnava.github.io/ANTs/} has outstanding non-rigid registration performances. We compare our FIRE model against SyN \cite{Avants2008} for non-rigid registration. Performances of affie registration are compared against the mutual information (MI) implemented in ANTs.

\subsection{Results and Discussion}
\begin{figure}[!t]
\centering
\includegraphics[width=0.9\linewidth]{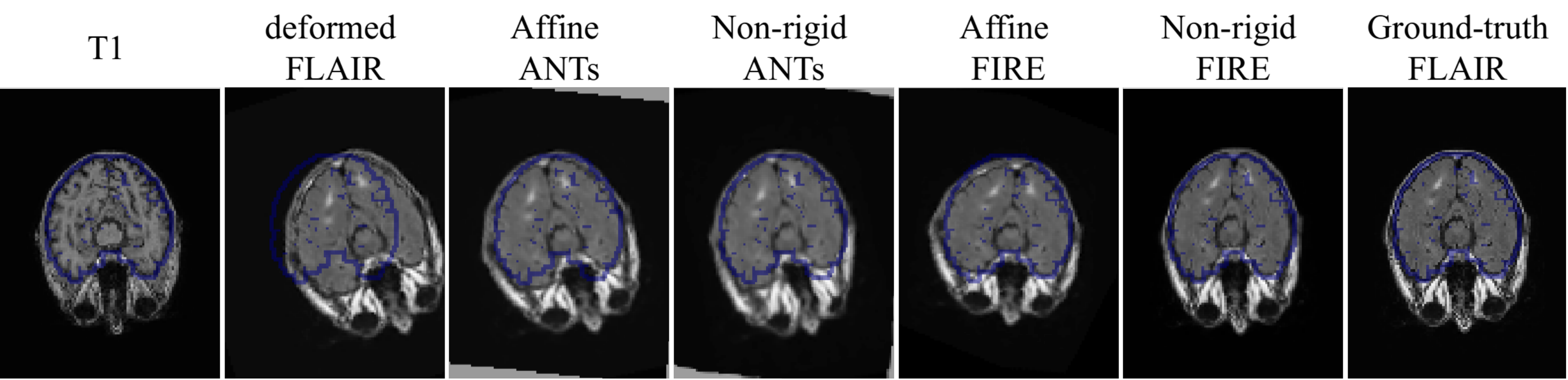}
\caption{Representative results of MRBrainS T1-FLAIR data.The outer contour of cerebrospinal fluid in the extracerebral space segmented on T1 images are shown in blue.}
\label{fig:MRBrainS_TF}
\end{figure}
Table \ref{tab:MRSResults} summarizes the Dice scores obtained from registration between MRBrainS T1 and FLAIR data, and Fig. \ref{fig:MRBrainS_TF} shows representative results. The proposed FIRE model achieved comparable results with SyN on the segmented cerebellum (Ce) and brain stem (BS), and higher scores on white matter (WHM). For 3D registration, our method obtained higher Dice scores on BS. In the example shown in Fig. \ref{fig:MRBrainS_TF}, FIRE achieved visibly better alignment between the outer contours of cerebrospinal fluid in the extracerebral space shown in blue. 
\begin{table*}[!t]
\caption{Results of 2D and 3D T1-FLAIR registration on MRBrainS data. Dice scores are calculated on cerebellum (Ce), white matter (WHM), brain stem (BS).}
\centering
\begin{tabular}{l@{\quad}c@{\quad}c@{\quad}c@{\quad}c@{\quad}c@{\quad}c}
\hline
Data& Object & unaligned & ANTs-affine & FIRE-affine & ANTs-SyN & FIRE \\
\hline
\hline
\multirow{3}{*}{\textbf{2D}}& BS & 11.62 (6.1) & 61.25 (3.7) & 62.90 (4.1) & 78.73 (7.3) & \textbf{80.68 (7.7)} \\
\cline{2-7}
& CE & 7.17 (4.4) & 63.32 (3.2) & 64.36 (4.0) & 75.72 (8.1) & \textbf{76.96 (7.3)} \\
\cline{2-7}
& WHM & 14.29 (7.5) & 59.12 (4.5) & 59.97 (4.4) & 81.36 (6.0) & \textbf{84.18 (3.7)} \\
\hline
\multirow{3}{*}{\textbf{3D}}& BS & 27.15 (9.2) & 67.15 (3.1) & 69.81 (4.1) & 79.77 (6.7) & \textbf{81.08 (7.0)} \\
\cline{2-7}
& CE & 28.38 (10.3) & 68.38 (3.6) & 70.62 (3.7) & 86.00 (6.9) & \textbf{86.13 (7.2)} \\
\cline{2-7}
& WHM & 20.27 (9.3) & 60.27 (3.8) & 60.61 (3.8) & 72.33 (7.4) & \textbf{72.56 (7.1)} \\
\hline
\end{tabular}
\label{tab:MRSResults}
\end{table*}

\begin{figure}[t]
\centering
\includegraphics[width=0.8\linewidth]{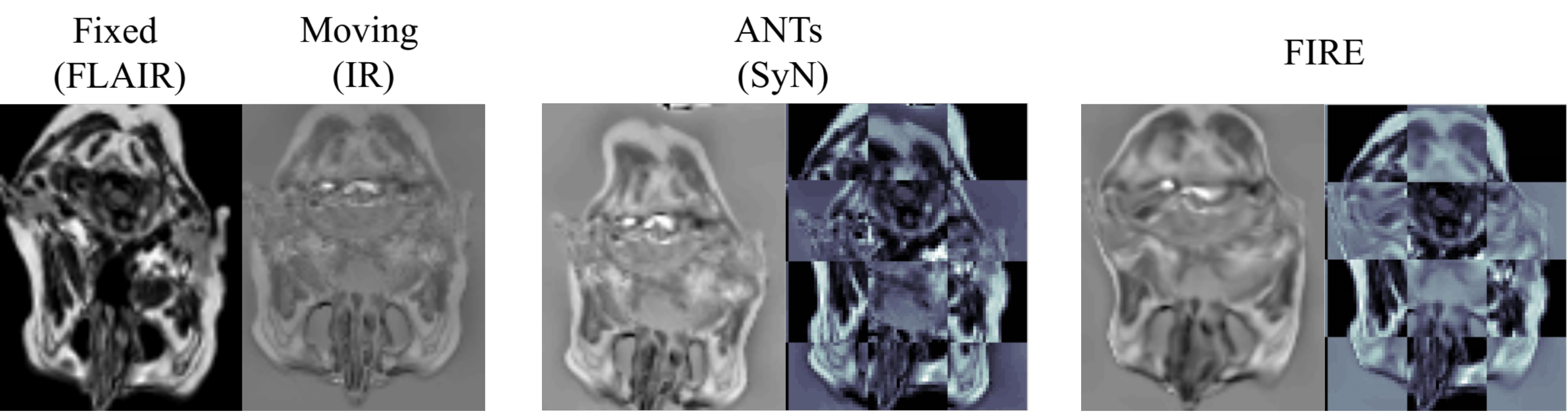}
\caption{Example results of registering the IR}
\label{fig:MRBrainS_IF}
\end{figure}

Registration between the IR and FLAIR images is difficult. We failed to produce a visible alignment using the SyN method implemented in ANTs after a grid search on its setup. As an example, the average dice score obtained on Ce using SyN and MI-based affine transformation is below 0.4 when the Dice score of the unaligned image is 0.6. Our method achieved a 0.69 Dice score for IR-FLAIR registration. An example of results is shown in Fig. \ref{fig:MRBrainS_IF}. 

\begin{figure}
\centering
\includegraphics[width=0.5\linewidth]{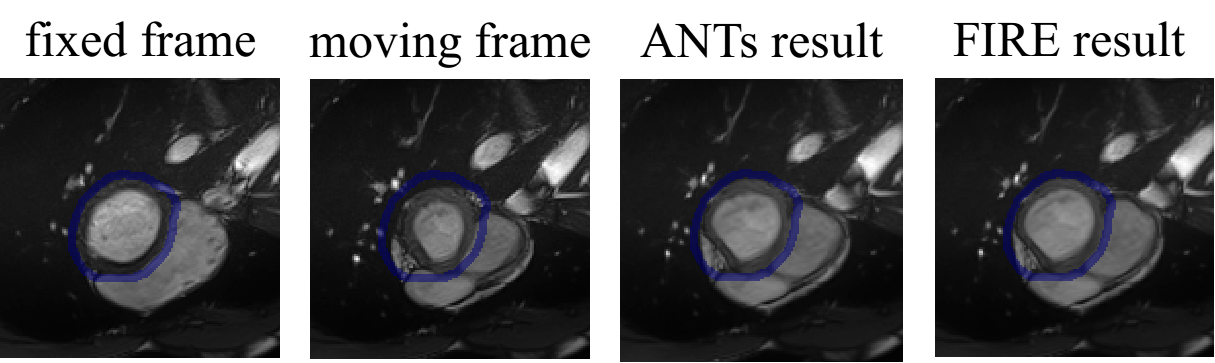}
\caption{Representative results of registration on ACDC data. Outer contours of myocardium are shown in blue.}
\label{fig:ACDCR}
\end{figure}

Table \ref{tab:ACDCResults} and Fig. \ref{fig:ACDCR} show the results of the inter-modality registration performed on the ACDC data. The data only show small local displacement between frames thus both compared methods Dice scores over 0.9 on LVe. The Fire model achieved comparable performances with SyN.

\begin{table*}[!t]
\caption{Results on ACDC data. Dice scores computed on left ventricular endocardium (LVe) and myocardium (Myo).}
\centering
\begin{tabular}{c@{\quad}c@{\quad}c@{\quad}c}
\hline
Object & unaligned & ANTs-SyN & FIRE \\
\hline
\hline
LVe & 65.75 (16.25) & \textbf{90.81 (4.3)} & 90.08 (5.5) \\
\hline
Myo & 51.97 (14.50) & 70.71 (5.6) & \textbf{71.66 (6.3)} \\
\hline
\end{tabular}
\label{tab:ACDCResults}
\end{table*}

\section{Conclusion}
We proposed a deep learning model which solves inverse-consistent inter- and intra-modality image registration problems through cross-domain synthesis. The new spatial transformation network and associated loss functions allow to predict both optimal affine and topology preserving non-rigid transformations. Experiments prove that our method have comparable state-of-the-art in both 3D and 2D registration tasks. We achieved better performances than the selected baseline on registration between IR and FLAIR brain data. The model has a new ``\begin{CJK*}{UTF8}{gbsn}火\end{CJK*}''-shape architecture formed by five sub-networks, thus we named it as FIRE. 

\bibliographystyle{splncs}
\bibliography{FireRegistration}

\begin{thebibliography}{10}

\bibitem{Rueckert2010}
Rueckert, D., Schnabel, J.A.:
\newblock Medical image registration.
\newblock In: Biomedical image processing.
\newblock Springer (2010)  131--154

\bibitem{Haskins2019}
Haskins, G., Kruger, U., Yan, P.:
\newblock Deep learning in medical image registration: A survey.
\newblock arXiv preprint arXiv:1903.02026 (2019)

\bibitem{Cao2017}
Cao, X., Yang, J., Zhang, J., Nie, D., Kim, M., Wang, Q., Shen, D.:
\newblock Deformable image registration based on similarity-steered cnn
  regression.
\newblock In: International Conference on Medical Image Computing and
  Computer-Assisted Intervention, Springer (2017)  300--308

\bibitem{Krebs2017}
Krebs, J., Mansi, T., Delingette, H., Zhang, L., Ghesu, F.C., Miao, S., Maier,
  A.K., Ayache, N., Liao, R., Kamen, A.:
\newblock Robust non-rigid registration through agent-based action learning.
\newblock In: International Conference on Medical Image Computing and
  Computer-Assisted Intervention, Springer (2017)  344--352

\bibitem{Rohe2017}
Roh{\'e}, M.M., Datar, M., Heimann, T., Sermesant, M., Pennec, X.:
\newblock Svf-net: learning deformable image registration using shape matching.
\newblock In: International Conference on Medical Image Computing and
  Computer-Assisted Intervention, Springer (2017)  266--274

\bibitem{Jaderberg2015}
Jaderberg, M., Simonyan, K., Zisserman, A.,  et~al.:
\newblock Spatial transformer networks.
\newblock In: Advances in neural information processing systems. (2015)
  2017--2025

\bibitem{Vos2019}
de~Vos, B.D., Berendsen, F.F., Viergever, M.A., Sokooti, H., Staring, M.,
  I{\v{s}}gum, I.:
\newblock A deep learning framework for unsupervised affine and deformable
  image registration.
\newblock Medical image analysis \textbf{52} (2019)  128--143

\bibitem{Christensen2001}
Christensen, G.E., Johnson, H.J.:
\newblock Consistent image registration.
\newblock IEEE transactions on medical imaging \textbf{20}(7) (2001)  568--582

\bibitem{Qin2019}
Qin, C., Shi, B., Liao, R., Mansi, T., Rueckert, D., Kamen, A.:
\newblock Unsupervised deformable registration for multi-modal images via
  disentangled representations.
\newblock arXiv preprint arXiv:1903.09331 (2019)

\bibitem{Zhang2018}
Zhang, J.:
\newblock Inverse-consistent deep networks for unsupervised deformable image
  registration.
\newblock arXiv preprint arXiv:1809.03443 (2018)

\bibitem{Avants2008}
Avants, B.B., Epstein, C.L., Grossman, M., Gee, J.C.:
\newblock Symmetric diffeomorphic image registration with cross-correlation:
  evaluating automated labeling of elderly and neurodegenerative brain.
\newblock Medical image analysis \textbf{12}(1) (2008)  26--41

\end{thebibliography}

\end{document}